\documentclass[conference]{IEEEtran}
\usepackage{cite}
\usepackage{graphicx}
\usepackage{amsmath}
\usepackage{amssymb}
\let\emptyset\varnothing
\usepackage[ruled,norelsize]{algorithm2e}
\SetKwInput{KwInput}{Input}
\usepackage[font=normalsize,labelfont=bf]{caption}
\SetKwInput{KwOutput}{Output}
\makeatletter
\newcommand{\removelatexerror}{\let\@latex@error\@gobble}
\makeatother

\usepackage[export]{adjustbox}
\graphicspath{ {images/} }

\begin{document}

%Here goes the title

\title{Adversarial Attacks on Face Detectors using Neural Net based Constrained Optimization}

%Authors List

\author
{\IEEEauthorblockN{Avishek Joey Bose}
\IEEEauthorblockA{Department of Electrical and\\Computer Engineering\\
University of Toronto\\
Email: joey.bose@mail.utoronto.ca }
\and
\IEEEauthorblockN{Parham Aarabi}
\IEEEauthorblockA{Department of Electrical and\\Computer Engineering\\
University of Toronto\\
Email: parham@ecf.utoronto.ca}
}
\maketitle

%Main body starts

\begin{abstract}
Adversarial attacks involve adding, small, often imperceptible, perturbations to inputs with the goal of getting a machine learning model to misclassifying them. While many different adversarial attack strategies have been proposed on image classification models, object detection pipelines have been much harder to break. In this paper, we propose a novel strategy to craft adversarial examples by solving a constrained optimization problem using an adversarial generator network. Our approach is fast and scalable, requiring only a forward pass through our trained generator network to craft an adversarial sample. Unlike in many attack strategies we show that the same trained generator is capable of attacking new images without explicitly optimizing on them. We evaluate our attack on a trained Faster R-CNN face detector on the cropped 300-W face dataset where we manage to reduce the number of detected faces to $0.5\%$ of all originally detected faces. In a different experiment, also on 300-W, we demonstrate the robustness of our attack to a JPEG compression based defense typical JPEG compression level of $75\%$ reduces the effectiveness of our attack from only $0.5\%$ of detected faces to a modest $5.0\%$.
\end{abstract}

\begin {IEEEkeywords}
Face Detection, Deep Learning, Adversarial Attacks,  Object Detection
\end{IEEEkeywords}

\section{Introduction}
\label{intro}
Artificial Intelligence and in particular deep learning has seen a resurgence in prominence, in part due to an increase in computational power provided by new GPU architectures. Consequently, deep neural networks have been applied to problems as varied as vehicle automation \cite{schmidhuber2015deep} and cancer detection \cite{esteva2017dermatologist}, making it imperative to better understand the ways in which these models are vulnerable to attack. In the domain of image recognition, Szegedy et al.\ \cite{szegedy2013intriguing} found that small, often imperceptible, perturbations can be added to images to fool a typical classification network into misclassifying them. Such perturbed images are called \textit{adversarial examples}. These adversarial examples can then be used in conducting \textit{adversarial attacks} on networks. There are several known methods for crafting adversarial examples, and they vary greatly with respect to complexity, computational cost, and the level of access required on the attacked model. 

In general, adversarial attacks can be grouped by the level of access they have to the attacked model and by their adversarial goal. \textit{White-box attacks} have full access to the architecture and parameters of the model that they are attacking; \textit{black-box} attacks only have access to the output of the attacked model \cite{papernot2017practical}. Adversarial attacks can also be grouped into \textit{targeted} and \textit{untargeted} attacks. Given an input image $x$, class label $y$ and a classifier $D(x): x \rightarrow y$ to attack, the goal of an untargeted attack is to solve $\textnormal{argmin}_{x'} L(x,x')$ such that $D(x) \neq D(x')$, where $L$ is a distance function between the unperturbed and perturbed inputs \cite{baluja2017adversarial}. The goal of a targeted attack is to solve $\textnormal{argmin}_{x'} L(x,x')$ such that $D(x') = t'$, where $t'$ is a target class chosen by the attacker, i.e. forcing an image of a cat to be classified as a dog by the model. 

A baseline approach is the Fast Gradient Sign Method (FGSM) \cite{goodfellow2014explaining}, where an attack is crafted based on the gradient of the input image, $x$, with respect to the classifier loss. FGSM is a white-box approach, as it requires access to the internals of the classifier being attacked. There are several strong adversarial attacks for attacking deep neural networks on image classification, such as L-BFGS \cite{szegedy2013intriguing}, Jacobian-based Saliency Map Attack (JSMA) \cite{papernot2016limitations}, DeepFool \cite{moosavi2016deepfool}, and Carlini-Wagner \cite{carlini2017towards} to name a few. However, these methods all involve some complex optimization over the space of possible perturbations, making them slow and computationally expensive. 

Compared to attacks on classification models attacking object detection pipelines are significantly harder. On state of the art object detectors like Faster R-CNN \cite{girshick2015fast} that use object proposals at different scales and positions before classifying them; the number of targets is orders of magnitude larger than classification models. In addition if the number of proposals attacked are a small subset of all total proposals the perturbed image may still be correctly detected with a different subset of proposals. Thus, a successful attack requires fooling all object proposals simultaneously. In this paper we show that it is possible to craft fast adversarial attacks on state of the art face detector. 

We propose a novel attack on a Faster R-CNN based face detector by producing small perturbations that when added to an input face image causes the pretrained face detector to fail. To create the adversarial perturbations we propose training a generator against a pretrained Faster R-CNN based face detector. Given an image, the generator produces a small perturbation that can be added to the image to fool the face detector. The face detector is trained offline only on unperturbed images and as such remains oblivious to the generator's presence. Over time, the generator learns to produce perturbations that can effectively fool the face detector it is trained with. Generating an adversarial example is fast and inexpensive, even more so than for FGSM, since creating a perturbation for an input only requires a forward pass once the generator is sufficiently well-trained. We validate the efficacy of our attack on the cropped 300-W test set \cite{belhumeur2013localizing} \cite{zhu2012face} \cite{le2012interactive} \cite{messer1999xm2vtsdb} \cite{phillips2005overview}. In a different experiment we test the robustness attack against a jpeg compression based defense as proposed in \cite{das2017keeping}\cite{dziugaite2016study} which we find helps only when the compression quality is low.

\section{Related Work}
\label{Related Work}
There are numerous adversarial attack strategies that have been proposed; in this paper we restrict our discussion to the ones that are closest to our attack. We direct the interested reader to this survey for a detailed description \cite{akhtar2018threat} of the different attack strategies and defenses. 
While adversarial attacks on classification networks have been widely studied object detection pipelines have been harder to attack \cite{xie2017adversarial}. This can largely be attributed to the fact that the number of targets per image for a detection net is much higher. That is to say given an image $x$ and a state of the art detection network \cite{girshick2015fast} which consists of a Region Proposal Network that proposes $N$ bounding boxes (typically in the thousands) which have high probability of containing an object that is then fed into a classification network to actually classify what the object class is. A successful attack in this setting thus consists of simultaneously fooling all $M$ bounding boxes. If $N=1$ then a object detection network is analogous to a classification network and as such the following attacks are relevant. 

\subsection{Fast Gradient Sign Method}

Given an image $x$, the Fast Gradient Sign Method (FGSM) \cite{goodfellow2014explaining} returns a perturbed input $x'$:
\begin{equation*}
x' = x - \epsilon \cdot \textnormal{sign} ( \nabla_x J (\theta, x, y))
\end{equation*}
where $J$ is the loss function for the attacked classifier and $\epsilon$ controls the extent of the perturbation, set to be sufficiently small that the perturbation is undetectable by eye. Intuitively, FGSM works by taking the gradient of the loss function to determine which direction a pixel's intensity should be changed to minimize the loss function. Then it shifts the pixel in the other direction. When done for all pixels simultaneously, the classifier is more likely to misclassify $x'$.
\subsection{Carlini-Wagner}

The Carlini-Wagner method \cite{carlini2017towards} is used for conducting both targeted and untargeted attacks. The adversarial goal is finding some minimal perturbation $\delta$ such that $D(x + \delta) = t'$, where $D$ is the classifier, $x$ is some input, $t'$ is the target class, and $\delta$ is the perturbation. This is expressed as: 
\begin{align*}
& \textnormal{argmin}_\delta \|\delta\|_p + c \cdot f(x + \delta) \\
& \textnormal{ s.t. } x + \delta \in {[0,1]}^n
\end{align*}
where $f$ is an objective function such that $D(x + \delta) = t' \Leftrightarrow f(x + \delta) \leq 0$. The Carlini-Wagner attack encourages the solver to find a perturbation such that the perturbed input will be classified as the target class $t$ with high confidence, at least relative to the other possible classes. The Carlini-Wagner attack is very strong -- achieving over 99.8\% misclassification on CIFAR-10 -- but is slow and computationally expensive \cite{carlini2017towards}.

\subsection{Adversarial Transformative Networks}

An Adversarial Transformative Network (ATN) is any neural network that, given an input image, returns an adversarial image to be used against a particular classifier(s). Baluja et al.\  provide a broad formulation \cite{baluja2017adversarial}:
\begin{equation*}
\textnormal{argmin}_\theta \sum_{x_i \in \mathcal{X}} \beta \cdot L_{\mathcal{X}} ( g_{f, \theta}(x_i), x_i ) + L_{\mathcal{Y}} ( f (g_{f, \theta} (x_i)), f(x_i) )
\end{equation*}
where $\beta$ is a scalar, $L_{\mathcal{X}}$ is a perceptual loss (e.g., the $L_2$ distance) between the original and perturbed inputs and $L_{\mathcal{Y}}$ is the loss between the classifier's predictions on the original inputs and the perturbed inputs. In the original paper, Baluja et al.\ \cite{baluja2017adversarial} use $L_{\mathcal{Y}} = L_2 (f(x'), r(f(x), t))$, where $r$ is a re-ranking function meant to encourage better reconstruction. ATNs were less effective than strong attacks like Carlini-Wagner, and the adversarial images they generated were not found to be transferable for use in black-box attacks. One key advantage ATNs have is that they are fast and inexpensive to use: an adversarial image can be created with just a forward pass through the ATN.

\subsection{Dense Adversary Generation}
The Dense Adversary Generation (DAG) \cite{xie2017adversarial} approach produces perturbations that are effective against object detection and semantic segmentation pipelines. The adversarial goal in DAG optimizes a loss function over multiple targets in an image. The target is a pixel or a
receptive field in segmentation, and object proposal in detection. The optimization process is done over multiple steps using gradient based methods; the stopping condition for DAG is either fooling all targets or until a maximum number of iterations reached.

\subsection{Overview of Faster R-CNN}
In this section we briefly review the Faster R-CNN architecture which builds upon predecessors R-CNN \cite{girshick2014rich} and Fast R-CNN \cite{girshick2015fast}. Faster R-CNN consists of a two stage detection pipeline which which is end to end differentiable. In the first stage is a Region Proposal Network (RPN) is a fully convolutional network for generating object
proposals at different scales and aspect ratios. To do this the authors introduce anchors of different scales and aspect ratios for each position of convolution. To account for proposed regions with different sizes due to the variability in anchors Region of Interest (ROI) pooling is used. ROI pooling transforms the different sized object proposals outputted by the RPN to the same size. The second stage of Faster R-CNN consists of a detector that refines the bounding box proposals from the RPN as well a classifier which identifies the class of each bounding box. The final output is constructed by thresholding the proposed boxes and using non-maximum suppression to reduce overlapping boxes.

\section{Problem Formulation}
Constructing adversarial examples for face detectors can be framed as a constrained optimization problem similar to the Carlini-Wagner attack.
\begin{align*}
& \textnormal{minimize} \; L(x,x + \delta) \\
& \textnormal{ s.t. }  D(x + \delta) = t' \\
& x + \delta \in {[0,1]}^n
\end{align*}
Here $L$ is a suitable norm such as $L_2$ that enforces similarity between the original and adversarial sample in input space. While $D$, $\delta$, and $t'$ are the trained face detector, generated perturbation, and background class for the detector respectively. This optimization problem is typically very difficult as the constraint $D(x + \delta) = t'$ is highly non-linear due to $D$ being a neural network. Instead, the problematic constraint can be moved to the objective function as a penalty term for violating the original constraint. Specifically, we ascribe a penalty for each of the targets that is correctly detected as a face in the adversarial sample. The reformulated problem can be stated as follows:

\begin{align*}
& \textnormal{minimize} \; L(x,x + \delta) + \lambda L_{\textnormal{misclassify}}(x+\delta) \\
& \textnormal{ s.t. } x + \delta \in {[0,1]}^n
\end{align*}

In this setup the nonlinear constraint is removed and added as a penalty with a constant $\lambda > 0 $ which balances the magnitude of the perturbation generated to the actual adversarial goal.

\section{Approach}
\label{Approach}
Optimizing over a single parameter per image is still difficult for a detection network. Intuitively, adversarial attacks against face detectors should perturb pixels largely in the face region of an image. Thus to construct a fast attack that can generalize to new instances we need to model the abstract concept of a face. Neural networks have been proven to be universal function approximators \cite{csaji2001approximation} with the flexibility of modeling abstract concepts in images \cite{olah2018building}. We generate a perturbation with a conditional generator network $G$ which can then be updated in tandem with the target model. $G$ produces a small perturbation that can be added to $x$ to produce an adversarial image $x'$. The face detector remains oblivious to the presence of $G$ while $G$'s loss depends on how well it can fool the face detector into misclassifying $x'$. Over time, $G$ produces perturbations that can effectively fool the face detector it is trained with. Once fully trained, $G$ can be used to generate image-conditional perturbations with a simple feed-forward operation. Crucially, having a neural network producing perturbations means that during test time creating an attack is at most a forward pass which is significantly faster than even the fastest classification attack, FGSM. Finally, this is a general attack as the optimization is done over all images in the dataset rather than on a per image basis allowing for generalization to new unseen instances without further optimization steps. 

\subsection{Threat Model}\label{ssec:threat}

\begin{figure*}[!htp]
  	\centering
  	\includegraphics[width=0.9\textwidth]{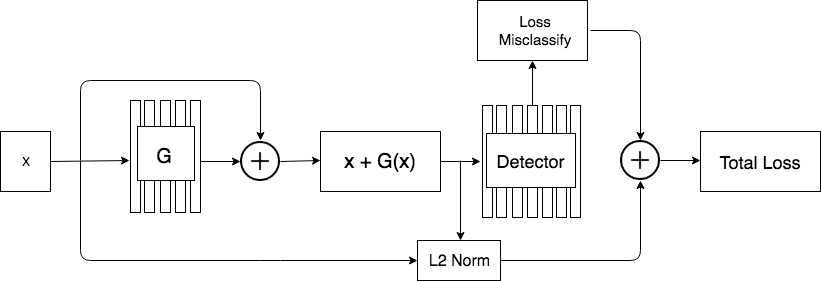}
    \caption{The proposed adversarial attack pipeline where a generator network $G$ creates image conditional perturbations in order to fool a face detector. G's loss is based on its success in fooling the face detector and the magnitude of the $L_2$ perturbation norm.}
    \label{fig:architecture}
\end{figure*}

Our model is most similar to that of an Adversarial Transformation Network (ATN) \cite{baluja2017adversarial}, a label that broadly applies to any generator network used to create adversarial attacks. However, it is significantly different from the specific type of ATN that was proposed and tested by Baluja et al.\ \cite{baluja2017adversarial}. Firstly, our attack is targeted against face detectors rather than purely image classifiers. We also train two networks a conditional generator $G$ using a pretrained detector over all targets proposed by the detector. In practice, we find that spending multiple iterations per image like DAG is crucial to effectively train G. Empirically we find that spending more time on a given example allows $G$ to generate perturbations that are smaller which when added to $x$ to produces an adversarial image $x'$ visually imperceptible to $x$. Throughout the training process the face detector remains oblivious to the presence of $G$ while $G$'s loss depends on how well it can fool the detector into misclassifying $x'$. Over time, $G$ produces perturbations that can effectively fool the face detector it is trained with. Once fully trained, $G$ can be used to generate image-conditional perturbations with a simple feed-forward operation. Fig \ref{fig:architecture} depicts the procedure for generating an adversarial example and the corresponding loss ascribed by the face detector as well an $L_2$ norm penalty to prevent large perturbations. We train $G$ end to end via gradient based optimization, backpropagating through the face detector network whose weights remain fixed while updating the weights of our generator network. 

\subsection{Learning the Generator}
The total loss on $G$ is a sum of $L_{misclassify}$ which forces $G$ to craft perturbations that lead to misclassification by the face detector and a $L_2$ norm cost between the original image $x$ and the adversarial sample $x'$. While there are many possible choices for $L_{misclassify}$, such as the likelihood of the perturbed images under the face detector we find that certain objectives much more robust to the choice of a suitable constant $\lambda > 0$. Typically, if $\lambda$ is very small i.e. $1e-4$ this results in adversarial samples that are almost identical to the original sample and thus are incapable of fooling the face detector. On the other hand with, if $\lambda$ is large i.e. $10$ this leads to images with large perturbations making them easily detectable visually by humans. Empirically, we find that choosing the same misclassification loss  as the Carlini Wagner attack is more robust to the choice of $\lambda$. Thus the total loss on $G$ for an input example is:
\begin{equation} \label{eq:Gloss}
L_G(x,x') = \Vert x - x' \Vert_2^2 + \lambda \sum_{i=1}^{N} \cdot (Z(x_i')_{\textnormal{background}} - Z(x_i')_{\textnormal{face}})^+
\end{equation}

Where $Z(x')$ is the unnormalized score of a specific class in object proposal $i$ out of $N$ total proposals on the perturbed image and $(x)^+$ denotes $\max(x,0)$. Like the attacks in DAG and DeepFool we find that it is necessary to perform multiple gradient steps on the same image, sometimes to convergence before optimizing for the next sample. The entire adversarial generator training procedure is illustrated in Algorithm 1.

\begin{figure}[!t]
 \removelatexerror
  \begin{algorithm}[H]
   \caption{Adversarial Generator Training}
   
  	\KwInput{
    	\begin{itemize}
    	\item[] Input Image $x$
        \item[] Face Detector $D$
        \item[] Generator $G$ with weights $\theta$
        \item[] Object Proposals $\Phi = \{ 1,2,..., N \}$
        \item[] Set of face labels for each object proposal $L_{face}$
        \item[] Maximum iteration $M$
        \item[] Perturbation Threshold $T$
        \item[] Step Size $\alpha$
    	\end{itemize}
        }
   \KwOutput{Adversarial Perturbation $\delta$}
   initialize $m = 0$, $\delta=0$, $L_2 = \infty$;\\
   \While{$L_{2} > T$ and $\Phi_m \neq \emptyset $ }{
   	 $\delta = G(x)$ \\
     $x' = \min(\max(x + \delta,1),-1)$ \\
     $Z(x') = D(x')$ \\
     $\Phi_m = {\textnormal{argmax}_c\{ {softmax(Z(x')} \} = L_{face}}$\\
     $L_{misclassify} = \sum_{i=1}^{N}(Z(x')_{\textnormal{background}} - Z(x')_{\textnormal{face}})^+$ \\
     $L_2 = \Vert x - x' \Vert_2^2$ \\
     $L_G(x,x') = L_2 + \lambda \cdot L_{\textnormal{misclassify}}$ \\
     $\theta = \theta - \alpha \nabla_{\theta} L_G(x,x') $\\
     	\If{$m > M$}{break \;}
    }
  \end{algorithm}
\end{figure}

\section{Experiments}
\label{section4}
We train the face detection model based on a pre-trained VGG16 \cite{simonyan2014very} model trained on the ImageNet dataset \cite{deng2009imagenet}. We randomly sample one
face image per batch for training. In order to fit it in the GPU
memory, the image is resized to a resolution of $600$ by $800$ pixels. For efficient training we restrict the number of object proposals to a maximum of $2000$ during training and $300$ during testing. In general we found that the Faster R-CNN face detector proposed many low confidence object proposals which led to a poor training signal for the generator. To fix this, we only consider object proposals for which the classifier probability is greater than $\alpha = 0.7 \%$, i.e. a $70\%$ detection threshold, while training. However, during testing we sweep through confidence values from $50 \%$ to $99\%$. We pretrain our Faster R-CNN face detector on the WIDER face dataset \cite{yang2016wider} for $14$ epochs using the ADAM optimizer with default settings \cite{kingma2014adam} before testing on the cropped 300 W dataset.

\subsection{Datasets}
The 300-W dataset, was first introduced for Automatic Facial Landmark
Detection in-the-Wild Challenge and is widely used as a benchmark for Face Alignment. Landmark annotations are provided following the Multi-PIE 68 points mark-up \cite{gross2010multi} and the 300-W test set consists of the re-annotated images from LFPW \cite{belhumeur2013localizing}, AFW \cite{zhu2012face}, HELEN \cite{le2012interactive}, XM2VTS \cite{messer1999xm2vtsdb} and FRGC \cite{phillips2005overview} datasets. Moreover, the 300-W test set is split into two categories, indoors and outdoors, of 300 images per category. In this paper we consider the cropped version of the combined indoor and outdoor splits of the 300-W dataset used for the IMAVIS competition.

\subsection{Semi-Whitebox attack}

\begin{figure*}[!htp]
\begin{center}
\includegraphics[width=0.265\textwidth]{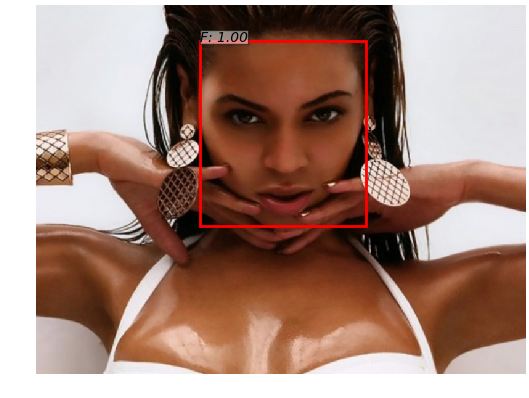}
\includegraphics[width=0.265\textwidth]{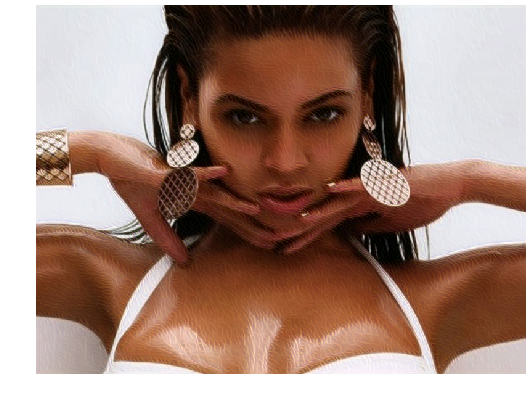}
\includegraphics[width=0.265\textwidth]{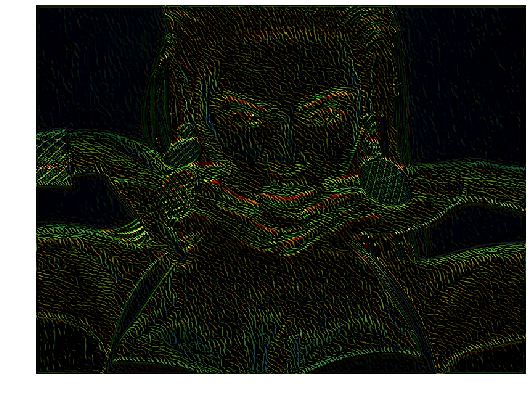}
\includegraphics[width=0.265\textwidth]{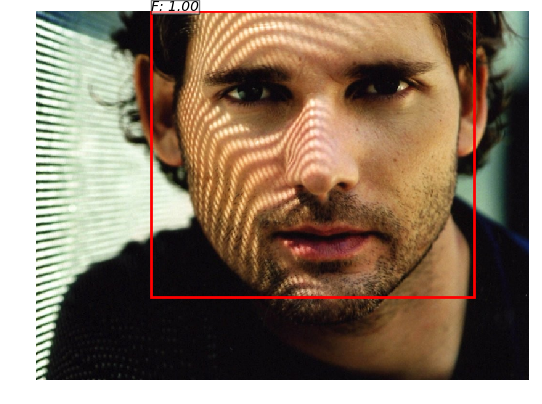}
\includegraphics[width=0.265\textwidth]{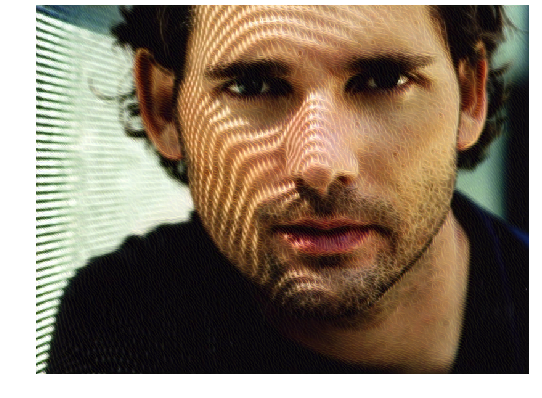}
\includegraphics[width=0.265\textwidth]{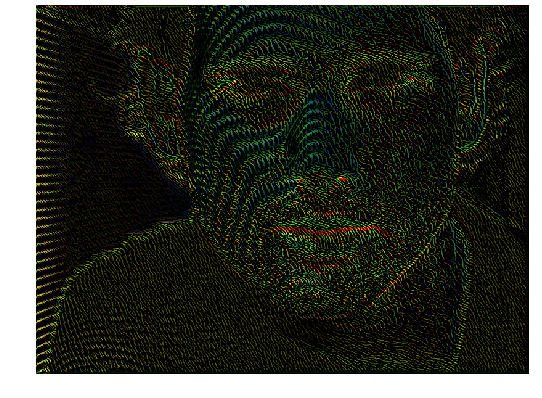}
\includegraphics[width=0.265\textwidth]{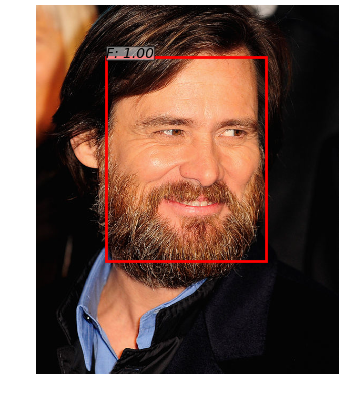}
\includegraphics[width=0.265\textwidth]{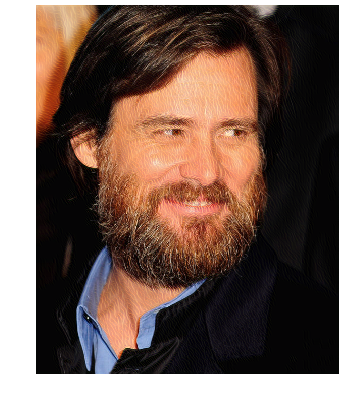}
\includegraphics[width=0.265\textwidth]{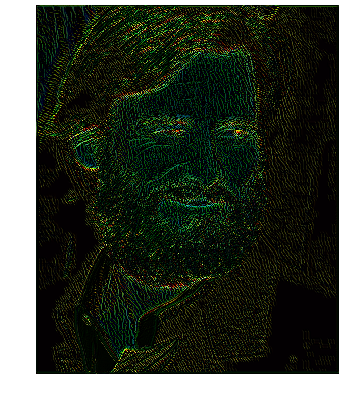}
\end{center}
\caption{A side by side comparison of face detections on 300-W dataset and the corresponding adversarial example with the generated perturbation that is not detected by the Faster R-CNN face detector. The detected face is enclosed in a bounding box with a corresponding confidence score of a face present. The perturbation was magnified by a factor of \textbf{$10$} to aid in their visualization }
\label{fig:onecol}
\end{figure*}

We classify our attack as somewhere between blackbox and whitebox as attacks crafted with a fully trained generator network do not require any internal information about the Faster R-CNN face detection model, but training the generator requires access to the face detector. We find that perturbations generated by our method is able to reduce the accuracy of the Faster R-CNN face detector from $99.5\%$ detected faces to just \textbf{$0.5\%$} on the cropped 300-W dataset. Furthermore, generating an adversarial perturbation is very fast as can be seen in Table \ref{tab:comptime}, a $45.2\%$ speed up over the Fast Gradient Sign Method and orders of magnitude faster than Carilini-Wagner. Fig \ref{fig:onecol} shows examples of the adversarial samples generated by our attack and the original image that was successfully detected by the face detector. Visually speaking the crafted adversarial samples have largely imperceptible differences but the generated perturbation is potent enough to reduce all object proposal scores below the detection threshold of $70\%$. To determine the impact of a specific detection threshold on the success of our attack we sweep through threshold values in the range of $50 \%$ to $99\%$, the results of which are presented in Table \ref{tab:alpha_label}. Indeed, our attack is robust to changes in detection thresholds and even when $\alpha=0.5$ we find that only $8$ faces are detected, a modest increase from $\alpha = 0.7$ which we fixed during entirety of training. We also find that our face detector is also fairly robust to changes in detection thresholds and $563$ faces are detected with $99\%$ confidence.    

\begin{table}[h]
\centering
\caption{Comparison of computation time different attack strategies for 1000 images on 1 Nvidia GTX-1080 Ti GPU.}
% with CPU Intel(R) Core(TM) i7-7700K CPU @ 4.20GHz.
\label{tab:comptime}
\begin{adjustbox}{width=0.7\linewidth}
\begin{tabular}{llll}
 & FGSM & C-W & Ours \\
\hline
\hline
Runtime & 2.21s & $>$6300s & \textbf{1.21s} \\
\hline
\end{tabular}
\end{adjustbox}
\end{table}

\begin{table}[ht]
\centering
\caption{Adversarial success rate given face detection confidence. The $\alpha$ value is the confidence threshold before an bounding box region is classified as a face. The columns represent the number of detected faces out of 600 faces.}
\begin{adjustbox}{width=0.7\linewidth}
\label{tab:alpha_label}
\begin{tabular}{lll}
 & Faster R-CNN & Our Attack \\
\hline
\hline
$\alpha=0.5$ & \qquad $599$ &  \qquad $8$ \\
$\alpha=0.6$ & \qquad $599$ &  \qquad $4$ \\
$\alpha=0.7$ & \qquad $597$ &  \qquad $3$ \\
$\alpha=0.8$ & \qquad $595$ &  \qquad $2$ \\
$\alpha=0.9$ & \qquad $593$ &  \qquad $1$ \\
$\alpha=0.99$ & \qquad $563$ &  \qquad $0$ \\
\hline
\end{tabular}
\end{adjustbox}

\end{table}

\subsection{Robustness to JPEG defense}
\begin{figure}[t]
\begin{center}
\includegraphics[height=0.7\linewidth,width=0.9\linewidth]{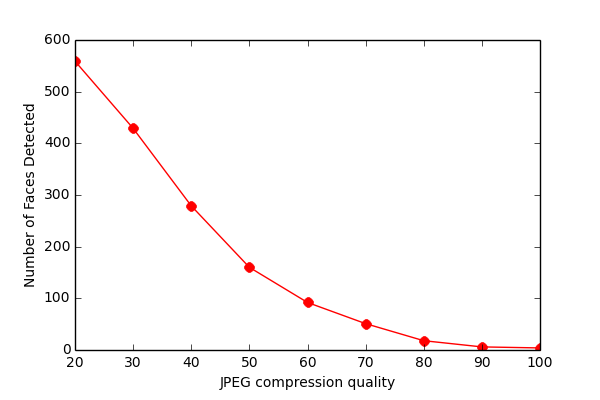}
\caption{The effect of JPEG compression on our adversarial attacks for Faster R-CNN face detectors.}
\label{fig:jpg_def}
\end{center}
\end{figure}

We evaluate the robustness of our attack under JPEG compression based defense which was shown to be effective against many of the attacks described in section II \cite{das2017keeping}. One theory for a JPEG based defense are that adversarial examples lie off the data manifold under which neural networks are so successful and by using JPEG compression the adversarial examples are projected back onto the data manifold removing their adversarial capabilities \cite{dziugaite2016study}. As can be seen in 
Fig \ref{fig:jpg_def} our attack is robust to JPEG compression when the quality is high but at very low levels the face detector is largely successful in detecting the face. A typical compression quality of $75\%$ yields a small increase in the fraction of detected faces from $0.5\%$ to $5\%$. 

\section{Conclusion}
\label {conclusion}
In this paper we introduce a novel adversarial attack on Faster R-CNN based face detectors by way of solving a constrained optimization problem using a generator network. Our attack is crafted through training a generator $G$ against a pretrained state of the art face detector based on the Faster R-CNN architecture. $G$ is responsible for learning to create fast image-conditional adversarial perturbations that can fool the face detector. Attacks crafted using $G$ can generalize to new face images without explicitly optimizing for them. We find that our attack is not only fast but also strong enough to fool the face detector on nearly every face image on the cropped 300-W dataset. Furthermore, the perturbations generated are strong enough to fool the face detector at low confidence levels. Finally, we show preliminary results of the robustness of our attack to a JPEG compression based defense strategy where image quality is not extremely poor.

\bibliographystyle{IEEEtran}
\bibliography{main}

% Generated by IEEEtran.bst, version: 1.12 (2007/01/11)
\begin{thebibliography}{10}
\providecommand{\url}[1]{#1}
\csname url@samestyle\endcsname
\providecommand{\newblock}{\relax}
\providecommand{\bibinfo}[2]{#2}
\providecommand{\BIBentrySTDinterwordspacing}{\spaceskip=0pt\relax}
\providecommand{\BIBentryALTinterwordstretchfactor}{4}
\providecommand{\BIBentryALTinterwordspacing}{\spaceskip=\fontdimen2\font plus
\BIBentryALTinterwordstretchfactor\fontdimen3\font minus
  \fontdimen4\font\relax}
\providecommand{\BIBforeignlanguage}[2]{{%
\expandafter\ifx\csname l@#1\endcsname\relax
\typeout{** WARNING: IEEEtran.bst: No hyphenation pattern has been}%
\typeout{** loaded for the language `#1'. Using the pattern for}%
\typeout{** the default language instead.}%
\else
\language=\csname l@#1\endcsname
\fi
#2}}
\providecommand{\BIBdecl}{\relax}
\BIBdecl

\bibitem{schmidhuber2015deep}
J.~Schmidhuber, ``Deep learning in neural networks: An overview,'' \emph{Neural
  networks}, vol.~61, pp. 85--117, 2015.

\bibitem{esteva2017dermatologist}
A.~Esteva, B.~Kuprel, R.~A. Novoa, J.~Ko, S.~M. Swetter, H.~M. Blau, and
  S.~Thrun, ``Dermatologist-level classification of skin cancer with deep
  neural networks,'' \emph{Nature}, vol. 542, no. 7639, pp. 115--118, 2017.

\bibitem{szegedy2013intriguing}
C.~Szegedy, W.~Zaremba, I.~Sutskever, J.~Bruna, D.~Erhan, I.~Goodfellow, and
  R.~Fergus, ``Intriguing properties of neural networks,'' \emph{arXiv preprint
  arXiv:1312.6199}, 2013.

\bibitem{papernot2017practical}
N.~Papernot, P.~McDaniel, I.~Goodfellow, S.~Jha, Z.~B. Celik, and A.~Swami,
  ``Practical black-box attacks against machine learning,'' in
  \emph{Proceedings of the 2017 ACM on Asia Conference on Computer and
  Communications Security}.\hskip 1em plus 0.5em minus 0.4em\relax ACM, 2017,
  pp. 506--519.

\bibitem{baluja2017adversarial}
S.~Baluja and I.~Fischer, ``Adversarial transformation networks: Learning to
  generate adversarial examples,'' \emph{arXiv preprint arXiv:1703.09387},
  2017.

\bibitem{goodfellow2014explaining}
I.~J. Goodfellow, J.~Shlens, and C.~Szegedy, ``Explaining and harnessing
  adversarial examples,'' \emph{arXiv preprint arXiv:1412.6572}, 2014.

\bibitem{papernot2016limitations}
N.~Papernot, P.~McDaniel, S.~Jha, M.~Fredrikson, Z.~B. Celik, and A.~Swami,
  ``The limitations of deep learning in adversarial settings,'' in
  \emph{Security and Privacy (EuroS\&P), 2016 IEEE European Symposium
  on}.\hskip 1em plus 0.5em minus 0.4em\relax IEEE, 2016, pp. 372--387.

\bibitem{moosavi2016deepfool}
S.-M. Moosavi-Dezfooli, A.~Fawzi, and P.~Frossard, ``Deepfool: a simple and
  accurate method to fool deep neural networks,'' in \emph{Proceedings of the
  IEEE Conference on Computer Vision and Pattern Recognition}, 2016, pp.
  2574--2582.

\bibitem{carlini2017towards}
N.~Carlini and D.~Wagner, ``Towards evaluating the robustness of neural
  networks,'' in \emph{Security and Privacy (SP), 2017 IEEE Symposium
  on}.\hskip 1em plus 0.5em minus 0.4em\relax IEEE, 2017, pp. 39--57.

\bibitem{girshick2015fast}
R.~Girshick, ``Fast r-cnn,'' \emph{arXiv preprint arXiv:1504.08083}, 2015.

\bibitem{belhumeur2013localizing}
P.~N. Belhumeur, D.~W. Jacobs, D.~J. Kriegman, and N.~Kumar, ``Localizing parts
  of faces using a consensus of exemplars,'' \emph{IEEE transactions on pattern
  analysis and machine intelligence}, vol.~35, no.~12, pp. 2930--2940, 2013.

\bibitem{zhu2012face}
X.~Zhu and D.~Ramanan, ``Face detection, pose estimation, and landmark
  localization in the wild,'' in \emph{Computer Vision and Pattern Recognition
  (CVPR), 2012 IEEE Conference on}.\hskip 1em plus 0.5em minus 0.4em\relax
  IEEE, 2012, pp. 2879--2886.

\bibitem{le2012interactive}
V.~Le, J.~Brandt, Z.~Lin, L.~Bourdev, and T.~S. Huang, ``Interactive facial
  feature localization,'' in \emph{European Conference on Computer
  Vision}.\hskip 1em plus 0.5em minus 0.4em\relax Springer, 2012, pp. 679--692.

\bibitem{messer1999xm2vtsdb}
K.~Messer, J.~Matas, J.~Kittler, J.~Luettin, and G.~Maitre, ``Xm2vtsdb: The
  extended m2vts database,'' in \emph{Second international conference on audio
  and video-based biometric person authentication}, vol. 964, 1999, pp.
  965--966.

\bibitem{phillips2005overview}
P.~J. Phillips, P.~J. Flynn, T.~Scruggs, K.~W. Bowyer, J.~Chang, K.~Hoffman,
  J.~Marques, J.~Min, and W.~Worek, ``Overview of the face recognition grand
  challenge,'' in \emph{Computer vision and pattern recognition, 2005. CVPR
  2005. IEEE computer society conference on}, vol.~1.\hskip 1em plus 0.5em
  minus 0.4em\relax IEEE, 2005, pp. 947--954.

\bibitem{das2017keeping}
N.~Das, M.~Shanbhogue, S.-T. Chen, F.~Hohman, L.~Chen, M.~E. Kounavis, and
  D.~H. Chau, ``Keeping the bad guys out: Protecting and vaccinating deep
  learning with jpeg compression,'' \emph{arXiv preprint arXiv:1705.02900},
  2017.

\bibitem{dziugaite2016study}
G.~K. Dziugaite, Z.~Ghahramani, and D.~M. Roy, ``A study of the effect of jpg
  compression on adversarial images,'' \emph{arXiv preprint arXiv:1608.00853},
  2016.

\bibitem{akhtar2018threat}
N.~Akhtar and A.~Mian, ``Threat of adversarial attacks on deep learning in
  computer vision: A survey,'' \emph{arXiv preprint arXiv:1801.00553}, 2018.

\bibitem{xie2017adversarial}
C.~Xie, J.~Wang, Z.~Zhang, Y.~Zhou, L.~Xie, and A.~Yuille, ``Adversarial
  examples for semantic segmentation and object detection,'' in
  \emph{International Conference on Computer Vision. IEEE}, 2017.

\bibitem{girshick2014rich}
R.~Girshick, J.~Donahue, T.~Darrell, and J.~Malik, ``Rich feature hierarchies
  for accurate object detection and semantic segmentation,'' in
  \emph{Proceedings of the IEEE conference on computer vision and pattern
  recognition}, 2014, pp. 580--587.

\bibitem{csaji2001approximation}
B.~C. Cs{\'a}ji, ``Approximation with artificial neural networks,''
  \emph{Faculty of Sciences, Etvs Lornd University, Hungary}, vol.~24, p.~48,
  2001.

\bibitem{olah2018building}
C.~Olah, A.~Satyanarayan, I.~Johnson, S.~Carter, L.~Schubert, K.~Ye, and
  A.~Mordvintsev, ``The building blocks of interpretability,'' \emph{Distill},
  vol.~3, no.~3, p. e10, 2018.

\bibitem{simonyan2014very}
K.~Simonyan and A.~Zisserman, ``Very deep convolutional networks for
  large-scale image recognition,'' \emph{arXiv preprint arXiv:1409.1556}, 2014.

\bibitem{deng2009imagenet}
J.~Deng, W.~Dong, R.~Socher, L.-J. Li, K.~Li, and L.~Fei-Fei, ``Imagenet: A
  large-scale hierarchical image database,'' in \emph{Computer Vision and
  Pattern Recognition, 2009. CVPR 2009. IEEE Conference on}.\hskip 1em plus
  0.5em minus 0.4em\relax IEEE, 2009, pp. 248--255.

\bibitem{yang2016wider}
S.~Yang, P.~Luo, C.-C. Loy, and X.~Tang, ``Wider face: A face detection
  benchmark,'' in \emph{Proceedings of the IEEE Conference on Computer Vision
  and Pattern Recognition}, 2016, pp. 5525--5533.

\bibitem{kingma2014adam}
D.~Kingma and J.~Ba, ``Adam: A method for stochastic optimization,''
  \emph{arXiv preprint arXiv:1412.6980}, 2014.

\bibitem{gross2010multi}
R.~Gross, I.~Matthews, J.~Cohn, T.~Kanade, and S.~Baker, ``Multi-pie,''
  \emph{Image and Vision Computing}, vol.~28, no.~5, pp. 807--813, 2010.

\end{thebibliography}

\end{document}